\documentclass[conference]{IEEEtran}
\IEEEoverridecommandlockouts
\usepackage{cite}
\usepackage{url}
\usepackage{amsmath,amssymb,amsfonts}
\usepackage{algorithmic}
\usepackage[linesnumbered,ruled]{algorithm2e}
\SetKwInput{KwInput}{Input}                % Set the Input
\SetKwInput{KwOutput}{Output}  
\SetKw{KwBy}{by}
\usepackage{graphicx}
\usepackage{textcomp}
\usepackage{xcolor}
\usepackage{textgreek}
\def\BibTeX{{\rm B\kern-.05em{\sc i\kern-.025em b}\kern-.08em
    T\kern-.1667em\lower.7ex\hbox{E}\kern-.125emX}}
\begin{document}

\title{Self Meta Pseudo Labels: Meta Pseudo Labels Without The Teacher\\

}

\author{\IEEEauthorblockN{1\textsuperscript{st} Kei-Sing Ng}
\IEEEauthorblockA{\textit{HKU Business School} \\
\textit{The University of Hong Kong}\\
Hong Kong, China \\
maxnghello@gmail.com}
\and
\IEEEauthorblockN{2\textsuperscript{nd} Qingchen Wang}
\IEEEauthorblockA{\textit{HKU Business School} \\
\textit{The University of Hong Kong}\\
Hong Kong, China \\
qcwang@hku.hk}
}

\maketitle

\begin{abstract}
We present Self Meta Pseudo Labels, a novel semi-supervised learning method similar to Meta Pseudo Labels \cite{b1} but without the teacher model. We introduce a novel way to use a single model for both generating pseudo labels and classification, allowing us to store only one model in memory instead of two. Our method attains similar performance to the Meta Pseudo Labels method while drastically reducing memory usage.
\end{abstract}


\section{Introduction}
Semi-supervised learning methods are essential techniques for many real-world machine learning problems as many companies have few and limited labeled data for model training. A recent research---Meta Pseudo Labels \cite{b1}improves the pseudo labeling method and has achieved new state-of-the-art performance on image classification problems \cite{b1, b2}. The method updates the teacher model based on the performance of the student model to generate better pseudo labels for training \cite{b1}. Meta Pseudo Labels works well with only 1,000 labeled examples and 60,300 unlabeled examples on the Street View House Numbers (SVHN) dataset \cite{b1, b3}, showing that semi-supervised learning can achieve better results than supervised learning.

Despite the strong performance of Meta Pseudo Labels, it requires storing two models in memory---a teacher model and a student model---during the training process. Some models can be prohibitively large with today's memory size, such as Efficient Nets \cite{b4} and GPT-3 model \cite{b5}, making it difficult to store two models in memory. Other semi-supervised learning approaches \cite{b6}can also achieve state-of-the-art performance using knowledge distillation \cite{b7}, but they also require storing a large and over-parameterized teacher
model or additional neural network layers in VRAM during training \cite{b8}. It is essential to find a way to reduce VRAM usage during training while still achieve strong performance. 

In this paper, we introduce a novel semi-supervised learning method as a variant of Meta Pseudo Labels. We design a mechanism for a model to learn from self-generated pseudo labels and improve quality of learning during the process. This model learns from both pseudo-labeled data and real labeled data.

\section{Related Works}

\subsection{Consistency Regularization}

There are many other semi-supervised learning approaches. Consistency regularization methods assume a model gives similar predictions for an unlabeled data sample and its perturbed version \cite{b2}. Given a neural network output of a data point, the objective of consistency regularization is to minimize the distance between it and the neural network output of its perturbed version. Some common distance measures include mean squared error and Kullback-Leiber divergence.

One example of consistency regularization is the Π-model by Samuli and Timo \cite{b9}. They use the stochastic nature of some neural network techniques to generate slightly different network outputs. For example, the dropout technique removes some neuron activation outputs randomly, making the final network output a stochastic variable. For every epoch of training, predictions are generated twice for each data example. The model is trained to minimize the difference between two outputs and the supervised loss on labeled data. 
 
 Another example is Unsupervised Data Augmentation (UDA) \cite{b10}. It improves the performance of a model by minimizing the consistency loss between predictions on original unlabeled examples and their noised versions. To generate the noised examples, UDA uses advanced data augmentations from supervised learning. It is shown to achieve an error rate of 4.20\% on the IMDb text dataset with only 20 labeled samples.

\subsection{Generative Models}

 Standard generative model (M1): Generative Models generate new examples from the input data distribution and learn the feature representations during training. This method transfers the learned feature representations to semi-supervised tasks and estimates the joint distribution of input data and labels. Because generative models do not require labels for input data during the training process, they can utilize a large amount of unlabeled data to learn transferable features for semi-supervised tasks. 
 
 Extended generative model (M2): During the training process of latent features in the M1 model, the labels of input data are not utilized. The training process of M2 model is similar to the one of M1 model but with an additional supervised loss if the labels are available. The labels are treated as latent variables if they are not available. 
 
 Stacked generative model (M1+M2): The stacked generative model method combines the M1 and M2 approaches. The method first trains the M1 model to learn the latent features. It then trains the M2 model using the learned latent features from the M1 model as a new representation of the input data. 
 
Variational Autoencoder for Semi-supervised Learning: 
Variational Autoencoder (VAE) \cite{b11} is a famous autoencoder model architecture. An autoencoder is a generative neural network model, with the objective of reconstructing the input data. It consists of an encoder and a decoder. The encoder projects the input data to a latent space, and the decoder reconstructs the input data from the latent vector. A VAE has an additional objective function to enforce that the latent vectors follow a unit Gaussian distribution. For a classification problem, an M2 VAE has an extra neural network classifier.
 
 Generative Adversarial Networks for Semi-supervised Learning: 
 A Generative Adversarial Network (GAN) \cite{b12} has a generator model and a discriminator
model. The generator model generates fake images, and is trained to generate images that are indistinguishable from the real images. A discriminator model takes both real and fake images as inputs, and its objective is to distinguish the fake data inputs from real ones. During the training process, the generator learns to generate better images, and the discriminator learns better representations of the input data. Some GANs such as BiGANs can generate highly realistic images \cite{b13}.

\section{Self Meta Pseudo Labels (SMPL)}

We present a novel method for semi-supervised learning. We mainly have two contributions:

\begin{itemize}
  \item We present a variant of Meta Pseudo Labels. Meta Pseudo Labels is currently the state-of-the-art pseudo labeling method. Our variant reduces VRAM usage by 19\% during training. 
  \item We introduce a novel way to train a model for semi-supervised learning, using a two-step gradient update. The second update is based on an evaluation of the performance of the model in the first update. 
\end{itemize}

To begin with, we first present the background of our method. 

\subsection{Background: Pseudo Labeling}
Pseudo labeling is a semi-supervised approach for deep neural networks. Conventional supervised deep neural network training methods update model parameters using a back-propagation algorithm with labeled data. We cannot utilize unlabeled data in vanilla supervised learning because the labels are not available. In a pseudo labeling approach, we have a teacher model and a student model. The teacher model generates pseudo labels for unlabeled data. The method then trains the student model with both labeled and unlabeled data simultaneously \cite{b14}. With pseudo labels, the method utilizes more unlabeled data during the training process to improve the final prediction result. 
One problem with conventional pseudo labeling is that the pre-trained teacher model is fixed during the training process. The student model may overfit on incorrect pseudo labels and results in confirmation bias \cite{b15}.

To address the problem, the teacher model of Meta Pseudo Labels is not fixed during the training process\cite{b1}. The teacher model receives feedback on the student model’s performance on the labeled examples, and is updated accordingly. The model learns to generate better pseudo labels, thus allowing the student model to converge better. This key modification makes Meta Pseudo Labels a new state-of-the-art with top-1 accuracy of 90.2\% on the ImageNet dataset \cite{b16}. 

\begin{figure*}[ht]
\centerline{\includegraphics[width=1\textwidth]{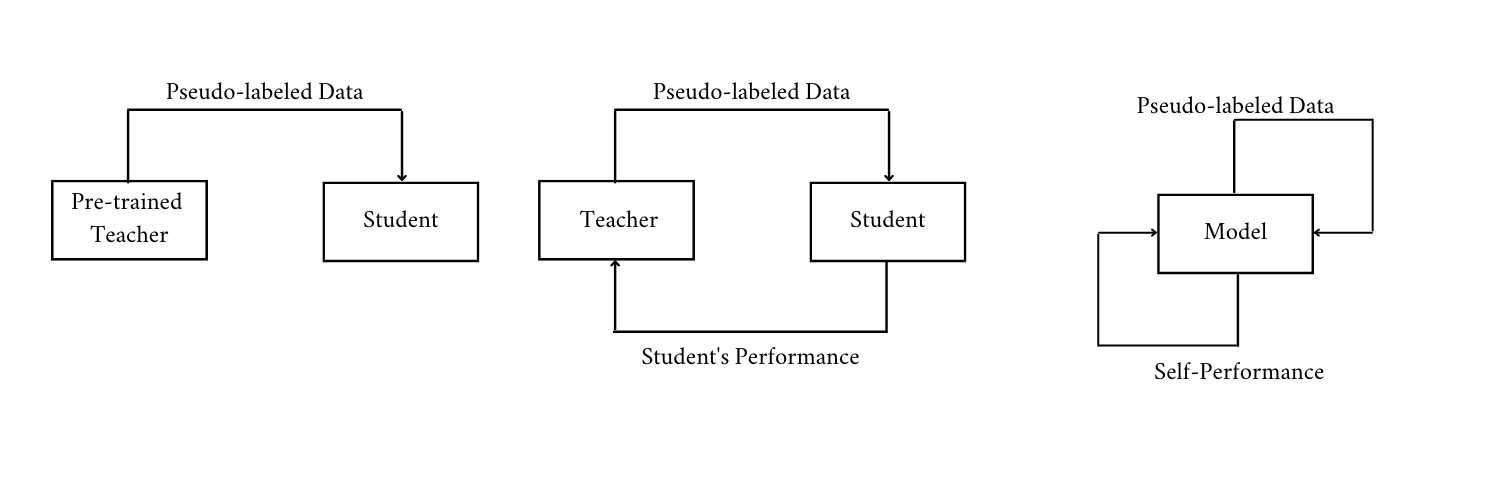}}
\caption{The difference between Pseudo Labels, Meta Pseudo Labels, and Self Meta Pseudo Labels. Left: Pseudo Labels, where the pre-trained teacher model is fixed to generate
pseudo labels. Middle: Meta Pseudo Labels, where the teacher model is trained and updated during training. The student model is trained with the pseudo labels generated by the teacher model. Right: Self Meta Pseudo Labels, where the teacher model and the student model are the same models. The model is updated based on its performance on labeled examples of the previous update.}
\label{fig}
\end{figure*}

\subsection{Self Meta Pseudo Labels} 
One drawback of Meta Pseudo Labels is using more VRAM than vanilla pseudo labeling. This method stores the teacher model in VRAM along with the student model during the training process. Our question is, do we need an external agent to generate the pseudo labels and evaluate the model? Human beings have the capability to act according to their hypotheses, observe, and adjust their actions. We want to simulate the process and update a model. We named it Self Meta Pseudo Labels.

 Unlike conventional pseudo labeling, we do not have an extra teacher model. To generate the pseudo labels, we pass a mini-batch of unlabeled data to the student model and get probabilistic predictions. We filter predictions of low confidence, and then convert the probabilistic predictions to hard pseudo labels. Every epoch contains two gradient updates. The first update is standard back-propagation using the hard pseudo labels \cite{b17}, and the second update depends on the performance of the first update:
 
 \begin{itemize}
  \item The first SGD: $ \theta^{'}_{M} = \theta_{M} - \eta_1 \cdot \nabla {L}_1(\theta_{M})$
  
  \item The second SGD: $ \theta^{''}_{M} = \theta^{'}_{M} - \eta_2 \cdot \nabla {L}_2(  \theta_{M} - \eta_1 \cdot \nabla {L}_1(\theta_{M}) )$
  
\end{itemize}

where $M$ is the neural network in Self Meta Pseudo Labels and $\theta_M$ is its parameters. 

We let ($x_l$, $y_l$) be a batch of labeled examples and their corresponding labels. We use $x_u$ to denote a batch of unlabeled examples. $x_{ua}$ is the augmented version of $x_u$. We use $p$ to denote the softmax prediction from the model and $H(p)$ to denote the hard labels.

\begin{algorithm}
\DontPrintSemicolon
  \KwInput{Labeled data $x_l$, $y_l$ and unlabeled data $x_u$.}
  {Initialize $\theta_M$.\;}
  \For{$k = 0$ \KwTo $N - 1$}{
    Sample a batch of unlabeled examples ($x_u$,$x_{ua}$) and a batch of labeled examples ($x_l$, $y_l$)\;
    Compute the hard pseudo labels $H(p_u$).\;
    Compute the loss $\mathcal{L}_1( \theta^{}_M )$ and gradient with the pseudo labels.\;
    Update the model: $ \theta^{'}_{M} = \theta_{M} - \eta_1 \cdot \nabla {L}_1(\theta_{M})$ \;
    
    Compute the new loss $\mathcal{L}_2( \theta^{'}_{M} )$ and gradient with the pseudo labels and ($x_l$, $y_l$).\;
    Update the model:
    $ \theta^{''}_{M} = \theta^{'}_{M} - \eta_2 \cdot \nabla {L}_2( \theta^{'}_{M} )$
    \;
    }
   \Return{$\theta_{M}$}

\caption{The Self Meta Pseudo Labels method}
\end{algorithm}

The first objective function $\mathcal{L}_1( \theta^{}_M )$ is a cross-entropy loss function $CE$ with labeling smoothing: 
\begin{equation}
\mathcal{L}_1( \theta^{}_M  )= CE(H(p_u), p_{ua})
\end{equation}

With label smoothing, we train the model to predict $(1-\alpha)$ instead of 1 for the correct class and $\alpha/(n-1)$ for the other classes where $\alpha$ is a small positive number and $n$ is the total number of classes.

The second objective function consists of two parts: 

\begin{equation}
\mathcal{L}_2( \theta^{'}_M  ) = \mathcal{L}_{UDA} + \lambda \mathcal{L}_{MPL}
\end{equation}

where $\mathcal{L}_{UDA}$ is the unsupervised data augmentation loss and $\mathcal{L}_{MPL}$ is the semi-supervised loss. $\lambda$ is a constant to control the ratio between the two terms. For the unsupervised data augmentation loss:

\begin{equation}
\mathcal{L}_{UDA}  = CE(y) +\beta_{k} E[{-p_ulog(p_{ua})])}
\end{equation}

\begin{equation}
\beta_{k} =  \beta_0 * Min(1, (k+1)/a)
\end{equation}

$\beta_{k}$ is a warm-up variable to control the magnitude. It gradually increases until the total number of steps reaches a constant $a$. $\mathcal{L}_{UDA}$ is masked and only predictions with high confidence are used.  $CE(y)$ is the cross-entropy loss for the labeled examples. For the semi-supervised loss: 

\begin{equation}
\mathcal{L}_{MPL} = \Delta CE * CE(H(p_{u}), p_u)
\end{equation}

We use $\mathcal{L}_{MPL}$ to evaluate the performance of the model after the first gradient update. $\mathcal{L}_{MPL}$ equals to the dot product $\Delta CE$ times the cross-entropy loss of $p_u$ and the hard pseudo labels. The dot product $\Delta CE$ is the difference in the cross-entropy loss of labeled examples before and after the first gradient update. In practice we subtract the moving average of $\Delta CE$ from $\Delta CE$ when we compute $\mathcal{L}_{MPL}$ to reduce the variance.

\subsection{Augmentation Strategies}

We use different data augmentation policies such as Unsupervised Data Augmentation, AutoAugment \cite{b18} and RandAugment \cite{b19} in our method to enhance the performance. We use UDA as an extended objective when training in the teacher role.

We combine data augmentation policies from AutoAugment and RandAugment in our method. AutoAugment is a method that automatically searches for combinations of data augmentation policies to improve the accuracy of a classification model. In every mini-batch, sub-policies are randomly chosen for each data example. It composes of many sub-policies such as translation, rotation, and shearing. It shows an improvement in accuracy on the CIFAR-10, CIFAR-100 \cite{b20}, SVHN, and ImageNet datasets. RandAugment is another data argumentation method that finds data argumentation policies with a reduced search space \cite{b19}. It removes the need for a separate search phase, and can be applied to various models and dataset sizes. We include a list of all the data augmentation policies used in our experiments in Appendix A.

\section{Experiments}

\subsection{Toy Experiment}

To better understand the method, we conduct a toy experiment on a small-scale dataset. We then compare the result between Self Meta Pseudo Labels and conventional supervised learning. 

\subsubsection{Dataset}

We use the moon dataset from Scikit-learn \cite{b21}. It is a simple toy dataset having 2d data points with two interleaving half circles on a 2d plane. We randomly generate 2,000 examples into two classes. We keep 6 examples as label examples randomly and use the rest as unlabeled data. The task of the experiment is to classify the examples correctly.

\subsubsection{Training details}

We remove all data augmentations and regularization losses. We use a simple neural network model with two fully connected hidden layers. Each layer has 8 units. We use a ReLU activation function \cite{b22}, and train the model with an initial learning rate of 0.1 for 1,000 steps. We use the last checkpoint as the final checkpoint and evaluate it on the whole dataset. For the supervised learning experiment, We keep the hyperparameters unchanged but we train the model with the labeled examples only. 

To better explain our approach, we illustrate the gradient descent process with an example. We project the cost function to a 3d space in Figure 2. The red arrow represents a gradient descent of vanilla supervised learning. It moves towards the global minimum. In SMPL, we have two gradient updates in every training epoch, represented with two blue arrows. The first update moves away from the global minimum while the second update corrects the direction. The final result is closer to the global minimum compared to vanilla supervised learning. 

\begin{figure}[t]
\centering
\includegraphics[width=0.5\textwidth]{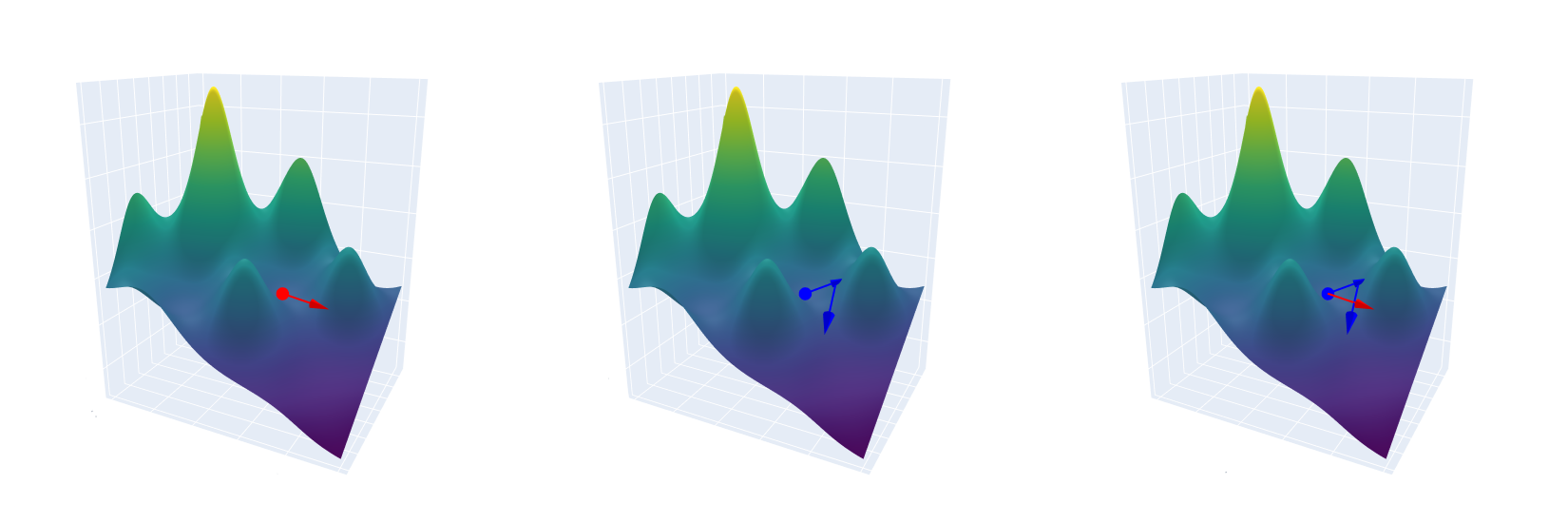}
\caption{\label{fig:gradient} Left: A conventional gradient descent (red arrow). Middle: A two-step gradient descent of Self Meta Pseudo Labels (blue arrows). Right: We illustrate the two gradient descents in one figure. }
\end{figure}

\subsubsection{Results}
We achieve 81.15\% and 76.55\% accuracies with Self Meta Pseudo Labels and the supervised learning respectively. We achieve 4.6\% more in accuracy with Self Meta Pseudo Labels than supervised learning using the same model infrastructure. Self Meta Pseudo Labels trains a better model by utilizing the information of unlabeled data during updates.  We visualize the result in Figure 3. 

\begin{figure*}[t]
\centering
\includegraphics[width=1\textwidth]{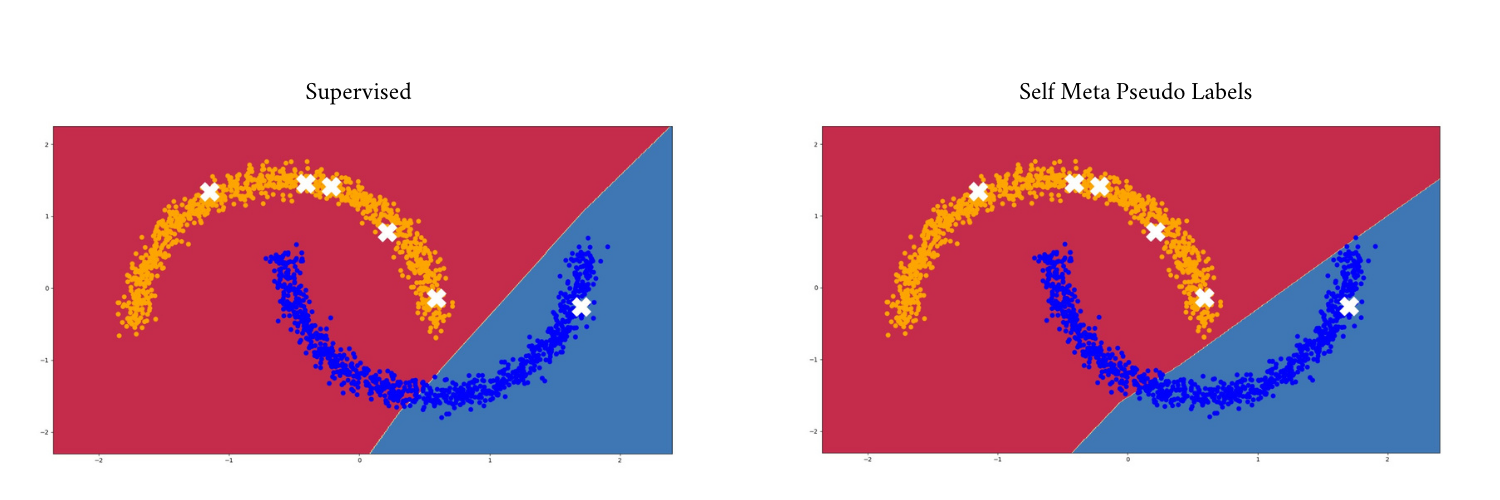}
\caption{\label{fig:moon} We conduct a toy experiment on the moon dataset. The result of supervised learning is on the left. The result of Self Meta Pseudo Labels is on the right. We color the two categories of examples into orange and dark blue. The labeled examples are marked in white. The model separates data points into red and blue.}
\end{figure*}

\subsection{CIFAR-10-4K, CIFAR100-10K, and SVHN-1K Experiments}

\subsubsection{Datasets}

We run experiments on three standard datasets: CIFAR-10-4K, CIFAR-100-10K and SVHN-1K. The CIFAR-10 and CIFAR-100 datasets contain 32x32 colour images in 10 classes and 100 classes respectively. The SVHN dataset contains images of digits from real-world house numbers photos. All three datasets contain less than 100k image samples which is comparable to many real-world machine learning problems. Many companies contain a limited amount of training samples and few labeled data. They require an efficient semi-supervised learning method. 

In our experiments, we keep a portion of labeled data and use the rest as unlabeled data. We keep 4,000 labeled images and use 46,000 unlabeled images in the CIFAR-10-4K dataset. For the CIFAR-100-10K dataset, we keep 10,000 labeled images, and use 40,000 images as unlabeled data for a total of 100 classes. For the SVHN-1K dataset, we keep 1,000 labeled images, and use about 603,000 images as unlabeled data for a total of 10 classes.

\subsubsection{Baselines}

Since we present a variant of Meta Pseudo Labels,
we directly compare the performance between Meta Pseudo Labels and our training method. We re-implement Meta Pseudo Labels, and use a WideResNet-28-2 \cite{b23} neural network model in our training. We compare the final results on the CIFAR-10-4K and SVHN-1K datasets. We use a same set of hyperparameter settings and augmentation methods for the two methods. The main difference between the two methods is we use two models in Meta Pseudo labels and one model in Self Meta Pseudo labels. We also train a WideResNet-28-2 neural network model on the CIFAR-100-10K dataset with Self Meta Pseudo Labels. 
		
For a fair comparison, we only compare Self Meta Pseudo Labels against methods that use the same model architectures. It is known that larger architectures can possibly improve any deep learning method's performance. Our method can also be used along with many other deep learning optimization techniques such as a different optimizer, neural architecture search, etc. 

\subsubsection{Training details}
Specifically, we have two stochastic gradient descent steps in every training epoch. In step one, we first draw a batch of labeled data ($x_l$, $y_l$) and a batch of unlabeled data $x_u$ stochastically. For every batch of unlabeled data $x_u$, we generate a batch of augmented version $x_{ua}$. We then generate hard pseudo labels using the model predictions of $x_l$, $x_u$ and $x_{ua}$. We compute the gradient and the first objective function $\mathcal{L}_1( \theta^{}_M )$. We also calculate the cross-entropy loss for ($x_l$, $y_l$) and save it for the step two computation. We update the model's parameters $\theta_M$ using a conventional stochastic gradient descent method. We use $ \theta^{'}_{M}$ to denote the updated parameters.

In step two, We generate new predictions $p{}'$ on $x_l$ and $x_u$ using the updated model. We then update the model based on the semi-supervised loss function $\mathcal{L}_{MPL}$ and the unsupervised data augmentation loss $\mathcal{L}_{UDA}$. We clip the gradient norm at 0.8. 

After the training phase, we finetune the best checkpoint on labeled data to improve the accuracy. In our finetuning process, we retrain the model with the labeled data, using stochastic gradient descent with a fixed learning rate of 5e-6 to update the model. We retrain the model for 8,000 epochs with a batch size of 512. Following the technique of Meta Pseudo Labels, we return the model at the final checkpoint because the number of labeled samples is limited for all three datasets.

\begin{table}[t]
\caption{The Hyper-parameters for Self Meta Pseudo Labels}
\begin{center}
\begin{tabular}{|c|c|c|c|}
\hline
\textbf{Hyper-parameter} & \textbf{\textit{CIFAR-10}}& \textbf{\textit{CIFAR-100}}& \textbf{\textit{SVHN}} \\
\hline
$\beta_0$  & 8 & 8 & 8 \\
\hline
 $a$  & 5,000 & 5,000 & 5,000 \\
 \hline
 $\alpha$  & 0.15 & 0.15 & 0.15 \\
 \hline
  Initial learning rate  & 0.05 & 0.05 & 0.0025 \\
 \hline
 Batch size   & 128 & 128 & 128 \\
 \hline
 Dropout rate on last layer   &  0.5 & 0.5  & 0.5 \\
\hline
\end{tabular}
\label{tab1}
\end{center}
\end{table}

\subsubsection{Results}

We were unable to successfully re-run the Meta Pseudo Labels experiments with the official released code and instructions \cite{b24}, see details in Appendix B. We replicate our version of Meta Pseudo Labels using Pytorch on the CIFAR-10-4K dataset and achieve an accuracy of 95.87\% compared to 96.11\% in the original paper, taking 0.41s for one training epoch and 6,861MiB VRAM on average. We achieve an accuracy of 95.91\% using Self Meta Pseudo Labels on the CIFAR-10-4K dataset with the same set of hyperparameters and setup. We spend 0.44s on one training epoch and use 5,537MiB VRAM on average, achieving a 19.3\% reduction in VRAM usage. We reduce one generation of pseudo labels for every training epoch compared to Meta Pseudo Labels. The training time is longer because, in the current version of our back-propagation implementation, it takes longer to keep the computation graph in memory after the first stochastic gradient descent step. On the SVHN-1K dataset, we achieve 94.55\% and 95.69\% accuracy with the Meta Pseudo Labels method and our method respectively, and also obtain a 19.1\% reduction in VRAM usage. We achieve an accuracy of 78.32\% on the CIFAR-100-4K dataset.

\begin{table}[t]
\caption{A comparison of testing accuracy on datasets}
\begin{center}
\begin{tabular}{|c|c|c|c|}
\hline
\textbf{Method} & \textbf{\textit{CIFAR-10}}& \textbf{\textit{CIFAR-100}}& \textbf{\textit{SVHN}} \\
\hline
 MixMatch \cite{b25} & 93.76\%  & 74.12\% &  96.73\% \\
\hline
 FixMatch(CTA) \cite{b26} & 95.69\%  & 76.82\% &  97.64\% \\
\hline
 SimPLE \cite{b27} & 94.95\%  & 78.11\% &  97.54\% \\
\hline
Meta Pseudo Labels  & 96.11\%  & N/A &  98.01\% \\
\hline
Meta Pseudo Labels & 95.87\%  & N/A & 94.55\%  \\ (our re-implemetation) & & &  \\
 \hline
 Self Meta Pseudo Labels  & 95.91\%  & 78.32\% & 95.69\%\\
\hline
\end{tabular}
\label{tab2}
\end{center}
\end{table}

\begin{table}[t]
\caption{A comparison of the VRAM usage on datasets}
\begin{center}
\begin{tabular}{|c|c|c|c|}
\hline
\textbf{Method} & \textbf{\textit{CIFAR-10}}& \textbf{\textit{CIFAR-100}}& \textbf{\textit{SVHN}} \\
\hline
Meta Pseudo Labels & 6,861MiB & N/A & 6,862MiB \\ (our re-implemetation) & & & \\
\hline
Self Meta Pseudo Labels  & 5,537MiB & 19,697MiB & 5,549MiB \\
\hline
\end{tabular}
\label{tab3}
\end{center}
\end{table}

\section{Conclusion}

In this paper we present a novel semi-supervised learning method as a variant of Meta Pseudo Labels that combines the teacher model and student model into one single model. We present a novel way to train a model with a two-step gradient update. The large reduction in VRAM usage compared with Meta Pseudo Labels is critical in solving many practice problems because many models are too large to fit in a single commercial GPU. Specifically, we can utilize a larger model with the reduction in VRAM usage under the same VRAM capacity. We believe the method can be applied to many vector inputs other than images.

\begin{appendices}

\section{Data Augmentation Policies}
Table IV is a list of data augmentation policies we use in the experiments. We refer readers to \cite{b18} for the detailed descriptions of these data augmentation policies.

\begin{table}[t]
\caption{Data augmentation policies that we used in our method}
\begin{center}
\begin{tabular}{|c|}
\hline
AutoContrast \\
\hline
Brightness \\
\hline
Color \\
\hline
Contrast \\
\hline
Cutout \\
\hline
Equalize \\
\hline
Invert \\
\hline
Sharpness \\
\hline
Posterize \\
\hline
Solarize \\
\hline
Rotate \\
\hline
ShearX \\
\hline
ShearY \\
\hline
TranslateX \\
\hline
TranslateY \\
\hline
\end{tabular}
\end{center}
\end{table}

\section{Re-running The Meta Pseudo Labels Experiments With The Official Code}
We use the code and instructions from the repository \url{https://github.com/google-research/google-research/tree/master/meta_pseudo_labels} \cite{b24}.

We create a variable for the project's ID in Google cloud shell.
\begin{verbatim}
export PROJECT_ID=project-id
\end{verbatim}

We configure the project and log into the instance.

\begin{verbatim}
gcloud config set project $PROJECT_ID
gcloud compute tpus execution-groups create \
--name=instance-1 \
--zone=us-central1-a \
--disk-size=300 \
--machine-type=n1-standard-16 \
--tf-version=2.6.0 \
--accelerator-type=v3-8

gcloud compute ssh instance-1 \
--zone=us-central1-a
\end{verbatim}

We go to the target folder and run the following command from the README.md. 

\begin{verbatim}
python -m main.py \
  --task_mode="train" \
  --dataset_name="cifar10_4000_mpl" \
  --output_dir="path/to/the/output/dir" \
  --model_type="wrn-28-2" \
  --log_every=100 \
  --master="path/to/the/tpu/worker" \
  --image_size=32 \
  --num_classes=10 \
  --optim_type="momentum" \
  --lr_decay_type="cosine" \
  --save_every=1000 \
  --use_bfloat16 \
  --use_tpu \
  --nouse_augment \
  --reset_output_dir \
  --eval_batch_size=64 \
  --alsologtostderr \
  --running_local_dev \
  --train_batch_size=128 \
  --uda_data=7 \
  --weight_decay=5e-4 \
  --num_train_steps=300000 \
  --augment_magnitude=16 \
  --batch_norm_batch_size=256 \
  --dense_dropout_rate=0.2 \
  --ema_decay=0.995 \
  --label_smoothing=0.15 \
  --mpl_student_lr_wait_steps=3000 \
  --uda_steps=5000 \
  --uda_temp=0.7 \
  --uda_threshold=0.6 \
  --uda_weight=8
\end{verbatim}

The program stops because of the missing class 'autocontrast'.
\begin{verbatim}
NameError: name 'autocontrast' is not defined
\end{verbatim}

\end{appendices}

\vspace{12pt}


\begin{thebibliography}{00}
\bibitem{b1}
Hieu Pham, Zihang Dai, Qizhe Xie, Minh-Thang Luong and Quoc V. Le.
\newblock Meta Pseudo Labels, 2020;
\newblock arXiv:2003.10580.
\bibitem{b2}
Yassine Ouali, Céline Hudelot and Myriam Tami.
\newblock An Overview of Deep Semi-Supervised Learning, 2020;
\newblock arXiv:2006.05278.
\bibitem{b3}
Yassine Ouali, Céline Hudelot and Myriam Tami.
\newblock The Street View House Numbers (SVHN) Dataset, 2011;
\newblock [Online]. Available:  
\newblock \url{http://ufldl.stanford.edu/housenumbers/}
\bibitem{b4}
Mingxing Tan and Quoc V. Le.
\newblock EfficientNet: Rethinking Model Scaling for Convolutional Neural Networks, 2019,
\newblock International Conference on Machine Learning, 2019;
\newblock arXiv:1905.11946.
\bibitem{b5}
Tom B. Brown, Benjamin Mann, Nick Ryder, Melanie Subbiah, Jared Kaplan, Prafulla Dhariwal, Arvind Neelakantan, Pranav Shyam, Girish Sastry, Amanda Askell, Sandhini Agarwal, Ariel Herbert-Voss, Gretchen Krueger, Tom Henighan, Rewon Child, Aditya Ramesh, Daniel M. Ziegler, Jeffrey Wu, Clemens Winter, Christopher Hesse, Mark Chen, Eric Sigler, Mateusz Litwin, Scott Gray, Benjamin Chess, Jack Clark, Christopher Berner, Sam McCandlish, Alec Radford, Ilya Sutskever and Dario Amodei.
\newblock Language Models are Few-Shot Learners, 2020;
\newblock arXiv:2005.14165.
\bibitem{b6}
Ting Chen, Simon Kornblith, Kevin Swersky, Mohammad Norouzi and Geoffrey Hinton.
\newblock Big Self-Supervised Models are Strong Semi-Supervised Learners, 2020;
\newblock arXiv:2006.10029.
\bibitem{b7}
Geoffrey Hinton, Oriol Vinyals and Jeff Dean.
\newblock Distilling the Knowledge in a Neural Network, 2015;
\newblock arXiv:1503.02531.
\bibitem{b8}
Linfeng Zhang, Jiebo Song, Anni Gao, Jingwei Chen, Chenglong Bao and Kaisheng Ma.
\newblock Be Your Own Teacher: Improve the Performance of Convolutional Neural Networks via Self Distillation, 2019;
\newblock arXiv:1905.08094.
\bibitem{b9}
Samuli Laine and Timo Aila.
\newblock Temporal Ensembling for Semi-Supervised Learning, 2016;
\newblock arXiv:1610.02242.
\bibitem{b10}
Qizhe Xie, Zihang Dai, Eduard Hovy, Minh-Thang Luong and Quoc V. Le.
\newblock Unsupervised Data Augmentation for Consistency Training, 2019;
\newblock arXiv:1904.12848.
\bibitem{b11}
Diederik P Kingma and Max Welling.
\newblock Auto-Encoding Variational Bayes, 2013;
\newblock arXiv:1312.6114.
\bibitem{b12}
Ian J. Goodfellow, Jean Pouget-Abadie, Mehdi Mirza, Bing Xu, David Warde-Farley, Sherjil Ozair, Aaron Courville and Yoshua Bengio.
\newblock Generative Adversarial Networks, 2014;
\newblock arXiv:1406.2661.
\bibitem{b13}
Jeff Donahue, Philipp Krähenbühl and Trevor Darrell.
\newblock Adversarial Feature Learning, 2016;
\newblock arXiv:1605.09782.

\bibitem{b14}
Dong-Hyun Lee.
\newblock Pseudo-label : The simple and efficient semi-supervised learning
  method for deep neural networks.
\newblock {\em ICML 2013 Workshop : Challenges in Representation Learning
  (WREPL)}, 07 2013.
  
  
\bibitem{b15}
Eric Arazo, Diego Ortego, Paul Albert, Noel E. O'Connor and Kevin McGuinness.
\newblock Pseudo-Labeling and Confirmation Bias in Deep Semi-Supervised Learning, 2019;
\newblock arXiv:1908.02983.


\bibitem{b16}
Olga Russakovsky, Jia Deng, Hao Su, Jonathan Krause, Sanjeev Satheesh, Sean Ma,
  Zhiheng Huang, Andrej Karpathy, Aditya Khosla, Michael Bernstein,
  Alexander C. Berg, and Li Fei-Fei.
\newblock Imagenet large scale visual recognition challenge, 2014.

\bibitem{b17}
Aram Galstyan and Paul~R Cohen.
\newblock Empirical comparison of “hard” and “soft” label propagation
  for relational classification.
\newblock International Conference on Inductive Logic Programming, pages
  98--111. Springer, 2007.

\bibitem{b18}
Ekin D. Cubuk, Barret Zoph, Dandelion Mane, Vijay Vasudevan and Quoc V. Le.
\newblock AutoAugment: Learning Augmentation Policies from Data, 2018;
\newblock arXiv:1805.09501.
\bibitem{b19}
Ekin D. Cubuk, Barret Zoph, Jonathon Shlens and Quoc V. Le.
\newblock RandAugment: Practical automated data augmentation with a reduced search space, 2019;
\newblock arXiv:1909.13719.
\bibitem{b20}
Alex Krizhevsky, Vinod Nair, and Geoffrey Hinton.
\newblock Cifar-10 and cifar-100 datasets.
\newblock [Online]. Available:  
\newblock \url{https://www.cs.toronto.edu/~kriz/cifar.html}



\bibitem{b21}
Sklearn Two Moon Dataset.
\newblock [Online]. Available:  
\newblock \url{https://scikit-learn.org/stable/modules/generated/sklearn.datasets.make_moons.html}
\bibitem{b22}
Abien Fred Agarap.
\newblock Deep Learning using Rectified Linear Units (ReLU), 2018;
\newblock arXiv:1803.08375.
\bibitem{b23}
Sergey Zagoruyko and Nikos Komodakis.
\newblock Wide Residual Networks, 2016;
\newblock arXiv:1605.07146.

\bibitem{b24}
\newblock Meta Pseudo Labels Code, 2021;
\newblock [Online]. Available: 
\newblock \url{https://github.com/google-research/google-research/tree/master/meta_pseudo_labels}

\bibitem{b25}
David Berthelot, Nicholas Carlini, Ian Goodfellow, Nicolas Papernot, Avital Oliver and Colin Raffel.
\newblock MixMatch: A Holistic Approach to Semi-Supervised Learning, 2019;
\newblock arXiv:1905.02249.

\bibitem{b26}
Kihyuk Sohn, David Berthelot, Chun-Liang Li, Zizhao Zhang, Nicholas Carlini, Ekin D. Cubuk, Alex Kurakin, Han Zhang and Colin Raffel.
\newblock FixMatch: Simplifying Semi-Supervised Learning with Consistency and Confidence, 2020;
\newblock arXiv:2001.07685.

\bibitem{b27}
Zijian Hu, Zhengyu Yang, Xuefeng Hu and Ram Nevatia.
\newblock SimPLE: Similar Pseudo Label Exploitation for Semi-Supervised Classification, 2021;
\newblock arXiv:2103.16725.
\newblock DOI: 10.1109/CVPR46437.2021.01485.

\end{thebibliography}
\end{document}